\documentclass[runningheads]{llncs}
\usepackage{graphicx}
\usepackage{comment}
\begin{document}
\title{KAS-term: Extracting Slovene Terms from Doctoral Theses via Supervised \\Machine Learning}
\titlerunning{KAS-term: Extracting Slovene Terms via Supervised Machine Learning}
%
\author{Nikola Ljube\v{s}i\'{c}\inst{1}\orcidID{0000-0001-7169-9152} \and
Darja Fi\v{s}er\inst{2,1}\orcidID{0000-0002-9956-1689} \and
Toma\v{z} Erjavec\inst{1}\orcidID{0000-0002-1560-4099}}

\authorrunning{N. Ljube\v{s}i\'{c} et al.}
%
\institute{Dept. of Knowledge Technologies, Jo\v{z}ef Stefan Institute, Ljubljana, Slovenia 
\email{\{nikola.ljubesic,tomaz.erjavec\}@ijs.si}
\and
Dept. of Translation, Faculty of Arts, University of Ljubljana, Slovenia
\email{darja.fiser@ff.uni-lj.si}}

\maketitle              
\begin{abstract}
This paper presents a dataset and supervised learning experiments for term extraction from Slovene academic texts. Term candidates in the dataset were extracted via morphosyntactic patterns and annotated for their termness by four annotators. Experiments on the dataset show that most co-occurrence statistics, applied after morphosyntactic patterns and a frequency threshold, perform close to random and that the results can be significantly improved by combining, with supervised machine learning,
all the seven statistic measures included in the dataset. On multi-word terms the model using all statistics obtains an AUC of 0.736 while the best single statistic produces only AUC 0.590. Among many additional candidate features, only adding multi-word morphosyntactic pattern information and length of the single-word term candidates achieves further improvements of the results.

\end{abstract}

\section{Introduction}

One of the cornerstones of academic language is its terminology, but for small languages, such as Slovene, it is unrealistic to expect that professional terminologists will  be able to fill the gaps for all the scientific fields still lacking a terminological dictionary and keep up with the rapid advancement of sciences and the new terms that are regularly coined in the process. The only viable solution is to enable the scientific communities to manage their own terminologies collaboratively, with a common infrastructure and some technical and linguistic support, by taking advantage of the growing number of scientific works available on-line in order to automatically extract terminologies and offer them as a starting point towards their consolidation by the community. In this paper we report on the first steps in this process for Slovene, by describing a manually annotated dataset of term candidates extracted via morphosyntactic patterns, performing analyses on the dataset and running a set of experiments on supervised learning of terminology extraction.

The rest of the this paper is structured as follows. Section \ref{sec:data} describes the corpus used for the dataset construction, the dataset annotation procedure and its encoding. Section \ref{sec:ana} analyses annotator agreement. Section \ref{sec:predic} gives the baseline experiments on predicting whether a term candidate is indeed a term, and Section \ref{sec:conc} gives conclusions and directions for further research.

\section{Related Work}

A broad overview of linguistic, statistical and hybrid approaches to automatic terminology extraction (ATE) is given in \cite{pazienza2005terminology}. Contemporary term recognition tasks are usually performed in two steps \cite{Nakagawa:Mori:03}: 1) candidate term extraction and 2) term scoring and ranking, which we follow in this paper as well. For scoring and ranking, we combine various statistical predictors in a supervised learning setting, inspired by \cite{loukachevitch2012automatic} who combine 16 features with logistic regression, which improves the best single result by removing 30-50\% of errors depending on the domain. Similarly, \cite{conrado2013machine} show on three domain corpora of Portuguese that combinations of 19 features significantly outperform well-known statistics for ATE.

Such approaches require ATE datasets, several of which are already available, most notably the ACL RD-TEC \cite{handschuh2014acl}, a dataset for evaluating the extraction and classification of computational linguistics terms. 
Reference datasets for terminology extraction in the biomedical domain are the GENIA corpus \cite{kim2003genia} and the CRAFT corpus \cite{bada2014concept}, where terms in the abstracts or scientific articles annotated with concepts from well-defined ontologies. 
In a reference dataset for the domain of automotive engineering \cite{Bernier-Colborne2014} the authors also apply annotation in running text, but allow for evaluation of extracted lists of term candidates. \cite{schafer2015evaluating} report surprisingly high inter-annotator agreement given the overall vagueness of the task of automatic term extraction without relying on an ontology. The authors generate a reference dataset on German DIY instructions and report a Fleiss $\kappa$ agreement among three annotators for multi-word terms of 0.59 and single-word terms of 0.61.

\section{The KAS-term Dataset} \label{sec:data}

\subsection{The Corpus}

The dataset was extracted from the KAS corpus of Slovene academic writing \cite{kas-jtdh} which was collected via the Open Science Slovenia aggregator \cite{openscience-ojster} harvesting the metadata of digital libraries of all Slovene universities, as well as other academic institutions. 
The KAS corpus contains, inter alia, 700 PhD theses (40 million tokens) from a large range of disciplines.\footnote{The complete KAS corpus and the KAS-Dr corpus of PhDs are available for exploring through the CLARIN.SI concordancers, \url{http://www.clarin.si/info/concordances/}.} For the term extraction experiments presented in this paper we focused on PhD theses from three fields: Chemistry, Computer Science, and Political Science, which we selected by matching them with their CERIF (Common European Research Information Format) keywords, thus obtaining 48 PhDs form Chemistry, 105 from Computer Science, and 23 from Political Science. We sampled 5 PhD theses per field for manual annotation, yielding all together 15 theses for further processing.

\subsection{Term Candidate Extraction}

From these three 15 PhD theses we first automatically extracted term candidates, using the CollTerm tool \cite{pinnis12-term} given a set of manually defined term patterns. These patterns were originally developed for the Sketch Engine terminology extraction module \cite{raslan-term}. For the present experiments we used only 31 nominal patterns, from unigrams and up to 4-grams, e.g. \verb!Nc.*,S.*,Nc.*,Nc.*g.*!. The identified term candidates were extracted in the form of lemma sequences and the most frequent inflected phrases, keeping those that appear at least three times in the corpus. The candidates were alphabetically sorted to remove bias stemming from frequency or statistical significance of co-occurrence, both provided by the CollTerm tool.

\subsection{Annotation Procedure}

We produced separate lists of term candidates for each doctoral thesis. Each of these lists was then annotated by four annotators. Annotators, who were graduate students of the three fields in focus, were asked to choose among one of the 5 labels:

\begin{itemize}
    \item{in-domain term: words and phrases that represent a term from the field in focus}
    \item{out-of-domain term: words and phrases that represent a term from a field other than the one in focus}
    \item{academic vocabulary: vocabulary that is typical of academic discourse}
    \item{irrelevant sequence: words and phrases that belong to general vocabulary, foreign-language expressions, definitions, fragments of terminology}
    \item{to be discussed: borderline cases that need to be discussed} 
\end{itemize}

Annotators were also given instructions how to deal with difficult cases, such as how to distinguish between terms and general or academic vocabulary, in- and out-of domain terms, term boundaries etc. To ensure maximum consistency of the annotations, annotation was performed in several cycles, each cycle covering candidates from a single thesis. At the end of each cycle the referee examined and discussed all the discrepancies among the annotators, as well as resolved the borderline cases (the cases annotated with the \emph{discuss} label).

\subsection{Dataset Encoding}

We constructed the final dataset with term candidates being our data instances, along with their metadata, manual annotations and some basic statistics. The final dataset consists of 22,950 instances. The metadata we encode are the thesis identifier, scientific field, annotation round, lemma and most frequent surface form sequence, morphosyntactic pattern and length in words. We encode all the four manual annotations in the dataset, preserving the information about the annotator (anonymized, 1-4) who performed a specific annotation. Finally, we encode seven statistics calculated with the CollTerm tool during the term candidate extraction. These statistics are the frequency of the term candidate, and its tf-idf, chi-square, dice, pointwise mutual information, and t-score values. We distribute the final dataset in comma-separated-value and json formats via the CLARIN.SI repository \cite{11356/1198}.\footnote{\url{http://hdl.handle.net/11356/1198}}

\section{Dataset Analysis}\label{sec:ana}

\subsection{Overall Analysis}

\begin{figure*}
\includegraphics[width=\textwidth]{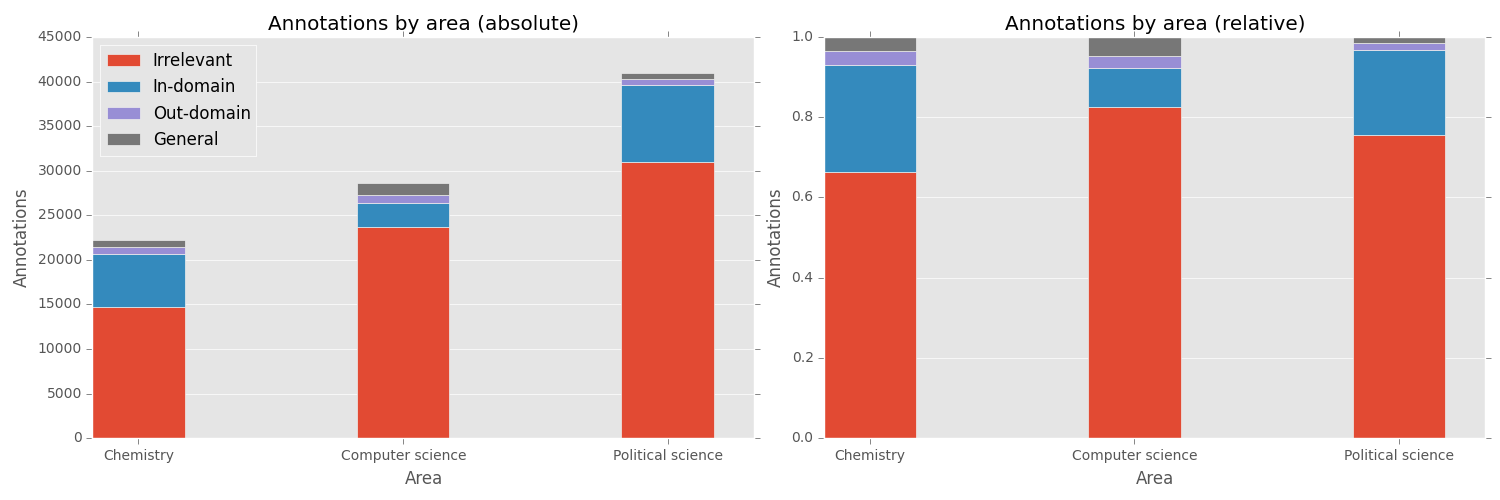}
\caption{\label{fig:area}Distribution of labels given the length of the term candidate (in tokens)}
\end{figure*}

The dataset consists of 22,950 instances (15,110 unique, 34\% duplicates), each covering a term candidate extracted from one of the 15 doctoral theses, 5 per each of the three areas covered. The two plots in Figure \ref{fig:area} show the distribution of term candidate through the three areas and the distribution of the annotations in each area, both on an absolute (left) and a relative scale (right). In that figure we can observe that most of the term candidates come from the area of political sciences, followed by computer science, which can be followed back to the length of the dissertations in each area. Regarding the term productivity of the three areas, computer science seems to have the highest percentage of irrelevant term candidates, while chemistry has the lowest. It is quite striking that out of the extracted term candidates, between 65 and 80 \% of term candidates can not be considered terms.




\subsection{Annotator Agreement}

Table \ref{tab:agreement} shows the observed agreements to be similar across all three areas, ranging from 72 to 78\%. However, Fleiss $\kappa$ suggests that the lowest agreement is in Computer science, followed by Political Science and  Chemistry. The only area in which moderate agreement ($>0.4$) is achieved is in Chemistry.

\begin{table}
\centering
\begin{tabular}{l|rrrr}
area & $\kappa$ & avg & min & max \\
\hline
\hline
Chemistry & 0.425 & 0.72 & 0.652 & 0.781 \\
Computer science & 0.282 & 0.781 & 0.744 & 0.835 \\
Political science & 0.341 & 0.748 & 0.658 & 0.832\\
\hline
Overall & 0.366 & 0.752 & 0.684 & 0.807 \\
\end{tabular}
\caption{\label{tab:agreement}Inter-annotator agreement in three areas, measured with Fleiss $\kappa$ and the average, minimum and maximum pairwise observed agreement.}
\end{table}

\section{Term Prediction} \label{sec:predic}

In this section we perform experiments on predicting whether a candidate is a term or not given the variables available in the prepared dataset. We perform all our experiments with \texttt{scikit-learn} \cite{scikit-learn}.

\subsection{Experimental Setup}

To be able to calculate a single, discrete gold label per instance in this dataset, we first map the four available categories to a binary schema. We consider two options: (1) considering only the \emph{in-domain} category as positive and the remaining as negative, we call this mapping \emph{exclusive} or (2) considering only the \emph{irrelevant} category as negative and the remaining as positive, calling this mapping \emph{inclusive}. While the exclusive mapping can be considered precision-oriented, the inclusive is more focused on recall. To inform our decision on which option to use in the remaining experiments, we calculate the overall Fleiss $\kappa$ for both. While the exclusive mapping obtains a $\kappa$ of 0.414 and an average observed agreement of 0.820, the inclusive mapping reaches a $\kappa$ of 0.369 and average observed agreement of 0.767. Given that we observe a higher agreement for the exclusive mapping, in the following ranking experiments presented in Section \ref{sec:ranking} we will primarily exploit that mapping, while for the final classification experiments between terms and non-terms presented in Section \ref{sec:classification}, we will investigate both of the mappings.


The explanatory variables we have at our disposal are frequency (\emph{freq}) and six co-occurrence statistics: chi-square (\emph{chisq}), Dice (\emph{dice}), pointwise mutual information (\emph{mi}), t-score (\emph{tscore}) and tf-idf (\emph{tfidf}). Due to its popularity, \emph{C-value} \cite{Frantzi2000} has been added to these experiments as well, although it uses information beyond frequency, such as intersection to other term candidates. We separate the prediction of multi-word terms (MWT) and single-word terms (SWT) as for single-word terms the only available variables are the frequency and the tf-idf statistic. For MWTs of all lengths all the seven variables are available.

\subsection{Single vs.\ Multiple Predictors in a Ranking Setting}
\label{sec:ranking}

\begin{figure}
\includegraphics[width=\columnwidth]{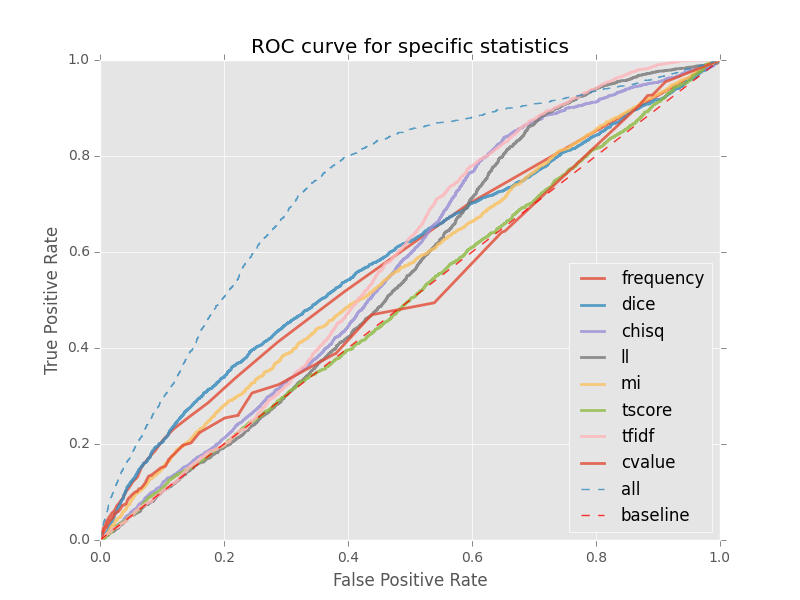}
\caption{\label{fig:roc}Receiver-operator-characteristic (ROC) curves for each of the variables, and for their combination \emph{all}. The \emph{baseline} is a random baseline.}
\end{figure}

We first evaluate the term predictability of each of the explanatory variables in isolation by performing ranking experiments.  As our response variable we use the rank of the term given the chosen statistic. Next we run an SVM regressor with an RBF kernel on all available variables, using the average annotation of each candidate (we have four annotations per candidate) as our response variable. For obtaining our predictions, we use cross-validation, in each iteration leaving one out of 15 available theses out for testing / annotation, and training on the remaining 14 doctoral theses. We evaluate each ranking by calculating the area under the curve (AUC) score which is a very convenient estimate as it does not require any decision on a threshold, i.e., the precision and recall trade-off. In Table \ref{tab:ranking} we give AUC scores for multi-word terms (MWT) and single-word terms (SWT) by each of the three areas for each of the statistics, as well for all statistics combined via the regressor, with the C-value (\emph{all+cv}) and without it (\emph{all})). In Figure \ref{fig:roc} we plot ROC curves for MWT for each statistic separately, as well as for the regressor combining all the available statistics (\emph{all}).

While performing experiments on MWT ranking via single variables, all the statistics seem to be similarly good rankers except for t-score which is almost at the level of random ranking (AUC of 0.5). It should be noted that the remaining statistics are also not that far from this random baseline, ranging from 0.52 to 0.59. When we combine the available variables into the regression model, leave-one-thesis-out-folding over the available data, we obtain single best results in all areas, as well as overall, the AUC climbing up to 0.736. Interestingly, the results are lowest in the Computer science area, and highest in Chemistry, i.e., they have the same order as in the Fleiss $\kappa$ inter-annotator agreement. It is hard to say whether the ranker has problems with specific areas due to the intrinsic complexity of the specific areas, or the lower annotation quality in these areas.

We further analyze the differences between specific statistics in the MWT setting via the ROC curves in Figure \ref{fig:roc}. These show the statistics to be easily divided into two groups: gradually peaking statistics (\emph{frequency}, \emph{dice} and \emph{mi}) and late peaking statistics (\emph{chisq}, \emph{ll} and \emph{tfidf}). While the gradually peaking statistics obtain best results in low true positive rates (TPRs) and false positive rates (FPRs), which is preferable in a precision-high setting, the late peaking ones take over at the bottom of the candidate list, i.e., in higher TPR and FPR values, therefore are more preferable in recall-oriented settings. However, the combination of the seven available statistics shows to obtain better results on the whole scale of TPR and FPR except for the final part of the ranking (FPR$>=$0.8 where late peakers \emph{ll} and \emph{tfidf} obtain slightly better results. This setting is, however, quite non-useful as it would require human inspection (or usage) of more than 80\% of the term candidates.

For single word terms (SWT), tf-idf is a much better ranker than frequency, the latter being very close to the random baseline. Combining the two available variables for SWTs, frequency and tf-idf, does not yield any improvements over ranking term candidates by the tf-idf statistic only, except for a similar result on the Chemistry area. Again, as with MWTs, the same order in the various ranking results between areas was obtained as is the Fleiss $\kappa$ inter-annotator agreement on the manual annotations.


\label{sec:auc}
\begin{table}
\centering
\begin{tabular}{l|rrrrr}
& Chemistry & Computer & Political & Overall \\
\hline
\hline
& \multicolumn{4}{c}{MWT}\\
\hline
\hline
freq & 0.593 & 0.580 & 0.596 & 0.586 \\
chisq & 0.582 & 0.537 & 0.605 & 0.571 \\
dice & 0.681 & 0.631 & 0.503 & 0.590 \\
ll & 0.590 & 0.560 & 0.566 & 0.557 \\
mi & 0.656 & 0.612 & 0.464 & 0.56 \\
tfidf & 0.544 & 0.538 & 0.613 & 0.582 \\
tscore & 0.457 & 0.454 & 0.585 & 0.505 \\
cvalue & 0.502 & 0.536 & 0.535 & 0.520 \\
\hline
all & \textbf{0.801} & \textbf{0.667} & \textbf{0.709} & \textbf{0.736} \\
all+cv & \textbf{0.801} & \textbf{0.655} & \textbf{0.711} & \textbf{0.736} \\

\hline\hline
& \multicolumn{4}{c}{SWT}\\
\hline
\hline
freq & 0.496 & 0.512 & 0.551 & 0.523 \\
tfidf & 0.791 & \textbf{0.673} & \textbf{0.71} & \textbf{0.703} \\
\hline
all & \textbf{0.8} & 0.509 & 0.687 & 0.613\\
\end{tabular}
\caption{\label{tab:ranking}Area under curve (AUC) results when exploiting single features and the combination of all features (\emph{all}). We discriminate between the three areas and the multi-word (MWT) and single-word terms (SWT).}
\end{table}



\subsection{Classification Experiments}
\label{sec:classification}
We move from ranking experiments to classification as this is our realistic final setting where, based on the content of each doctoral thesis, or any other scientific writing, we generate a manageable list of terms. During the classification experiments we continue differentiating between multi-word terms (MWTs) and single-word terms (SWTs) as the two cases have a very different amount of explanatory information available. While we have 7 statistics available for MWTs, there are only two available in the SWT case.

\subsubsection{Mapping Comparison}
\label{sec:mapping}
Our first experiment on MWTs uses an SVM classifier with an RBF kernel, reporting precision, recall and F1 on the term class obtained via leave-one-thesis-out cross-validation, on each area separately, and all together. We perform an experiment on each of the response variable mappings discussed at the beginning of the section, the exclusive mapping (where only the \emph{in-domain} label is considered to be positive) and the inclusive one (where only the \emph{irrelevant} label is considered to be negative). The results are reported in Table \ref{tab:mwt}. Again, the results per area depict that on both mappings the classification is the simplest in Chemistry where annotators agree among themselves best, followed by Political Science and Computer Science.

Interestingly, the inclusive mapping gives more precise results, 
with a small loss in recall, 
obtaining an overall significantly better F1 of five points. Given that the annotator agreement was better on the exclusive than the inclusive mapping, the only explanation for this difference in performance is to be sought in our explanatory variables, i.e., the corpus co-occurrence statistics. It seems that these statistics differentiate better between non-terms (\emph{irrelevant label}) and any kind of terms, being in the same (\emph{in-domain}), another (\emph{out-domain}), or general science domain (\emph{general}), than between in-domain terms and the rest. Given these results, the experiments on improving the classifier will be focused on the inclusive mapping as (1) we consider the output of both mappings to be different, but equally relevant for terminologists and (2) we seek to obtain overall best possible classification performance for an as-clean-as-possible output to be given to terminologists and other interested parties. On the inclusive mapping the results in Chemistry are very encouraging, with 56\% of the positive labeled candidates being terms, and among those, 79\% of all terms being present. Overall these numbers are lower, with 41\% of candidates in the positive class being terms, and two out of three terms being labeled with the positive class.

\begin{figure}
\includegraphics[width=\columnwidth]{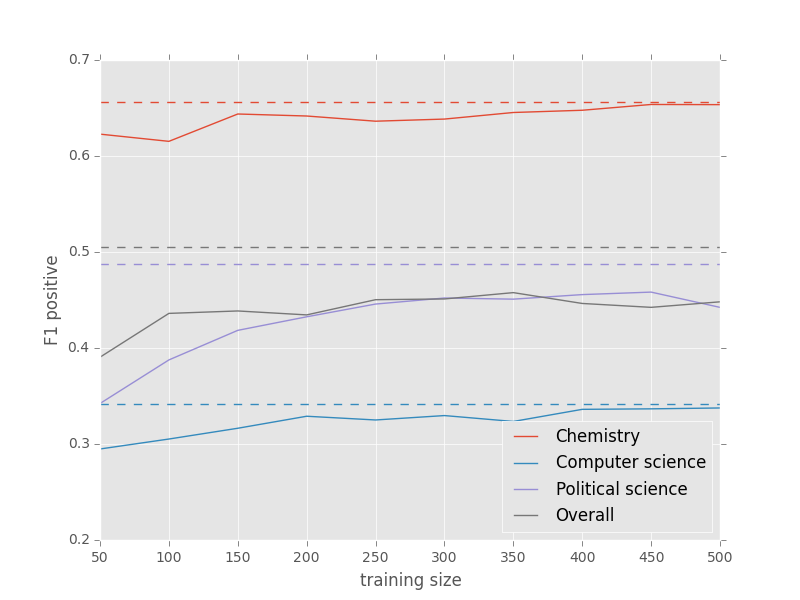}
\caption{\label{fig:learning}Learning curves per area. The reported results are average F1 on positive class. The label mapping is inclusive. The dashed lines represent the result obtained on full datasets.}
\end{figure}

\begin{table}
\centering
\begin{tabular}{l|rrr}
& P & R & F1 \\
\hline\hline
& \multicolumn{3}{c}{exclusive mapping} \\
\hline\hline
Chemistry & 0.453 & 0.827 & 0.586 \\
Computer Science & 0.148 & 0.777 & 0.249 \\
Political Science & 0.361 & 0.651 & 0.464 \\
Overall & 0.339 & 0.690 & 0.455 \\
\hline\hline
& \multicolumn{3}{c}{inclusive mapping} \\
\hline\hline
Chemistry & 0.564 & 0.783 & 0.656 \\
Computer Science & 0.221 & 0.752 & 0.342 \\
Political Science & 0.391 & 0.649 & 0.488 \\
Overall & 0.409 & 0.661 & 0.505 \\

\end{tabular}
\caption{\label{tab:mwt}MWT classification results (precision, recall, F1) on the positive class for the three areas and overall, both for the exclusive and inclusive annotation mappings.}
\end{table}

\subsubsection{Sensitivity to Amount of Training Data}


To measure the sensitivity of the task to the amount of training data, we calculated learning curves for MWTs only while applying the inclusive mapping. The resulting learning curves are presented in Figure \ref{fig:learning}. They show for the learning to be quite steady with rather flat learning curves already at the level of having just a few hundred annotated term candidates. The only area for which more annotations are useful, regardless of the rather flat learning curve, is Political science.


\subsubsection{Additional Features}

We wrap up our classification experiments by investigating additional features that are easily extractable from the KAS-term dataset. We also investigate whether any useful information from the context of terms and non-terms can be obtained from the full texts of doctoral theses. 
We first investigate whether \emph{Oversampling} instances for which the annotators were in agreement
boosts our performance. We next add the \emph{Candidate length} in form of one-hot encodings, investigating whether features behave differently depending on the length of the candidate. Next we add \emph{Average token length} of the candidate with the intuition that very short tokens probably do not form good term candidates. We continue with the most promising feature, the morphosyntactic \emph{Pattern} the candidate satisfies. It is to expect that some patterns produce better candidates than other. We finish with the \emph{Context} feature which presents the certainty of a context-based classifier for the candidate to be a term. The context-based classifier is an SVM with a linear kernel, features of the classifier being frequencies of tokens occurring in a 3-token window around all the occurrences of a term candidate in the respective doctoral thesis. 


The results on MWT candidate classification are presented in Table \ref{tab:mwtextra} showing that none of the examined features improves the overall results except the pattern feature as, naturally, some patterns produce better term candidates than others. This information yields a 2-point improvement on the overall F1 score.

\begin{table}
\centering
\begin{tabular}{l|rrr}
& P & R & F1 \\
\hline\hline
Original & 0.406 & 0.660 & 0.503 \\
Oversampling & 0.417 & 0.635 & 0.503 \\
Candidate length & 0.407 & 0.646 & 0.500 \\
Average token length & 0.375 & 0.742 & 0.498 \\
Pattern & 0.392 & 0.776 & \textbf{0.521}\\
Context & 0.399 & 0.688 & 0.505\\
\end{tabular}
\caption{\label{tab:mwtextra}MWT classification results (precision, recall, F1) on the positive class as additional features are added.}
\end{table}

Our SWT experiments are based on using \emph{tf-idf} as our only feature and, to double-check the negative ranking result for frequency from Section \ref{sec:ranking}, we add \emph{Frequency} as our first additional feature to be examined. We follow with adding the remaining features as with MWTs, except that we omit \emph{Candidate length} and \emph{Patterns}, as SWTs are all of same length and based on one single pattern.

As expected, the results in Table \ref{tab:swt} show that the \emph{Frequency} feature deteriorates the results, similarly as in the ranking task, while \emph{Oversampling} has a similar impact as with MWTs, improving precision and lowering recall, with no significant difference in F1. The \emph{Context} feature does not change the result significantly, similar as in the MWT setting. Finally, the feature that we expected to improve results, \emph{Average token length}, did actually improve F1 by 2.5 points.

\begin{table}
\centering
\begin{tabular}{l|rrr}
& P & R & F1 \\
\hline\hline
Original & 0.405 & 0.563 & 0.471 \\
Frequency & 0.417 & 0.522 & 0.463 \\
Oversampling & 0.408 & 0.557 & 0.471 \\
Average token length & 0.422 & 0.601 & \textbf{0.496} \\
Context & 0.427 & 0.510 & 0.465 \\
\end{tabular}
\caption{\label{tab:swt}SWT classification results (precision, recall, F1) on the positive class as additional features are added.}
\end{table}
\vspace{-0.6cm}

\section{Conclusion} \label{sec:conc}

In this paper we presented the dataset and a machine-learning approach for automatic terminology extraction from Slovene academic texts from three different scientific areas. The obtained Fleiss $\kappa$ coefficient classifies the agreement between annotators to be only fair, depicting the high complexity of the annotation task. 
We then analysed term predictability of various statistics included in the dataset, showing that
when combining all the available statistics, we obtain a significant improvement on multiword terms, with a relative improvement of AUC of 25\% over the single best-performing statistic. We further improve our multi-word term predictor by adding the information on the morphosyntactic pattern and our single-word term predictor via the character length of the term candidate. Interestingly, adding the context information does not improve our results on either of the problems.




\section*{Acknowledgements}
The work described in this paper was funded by the Slovenian Research Agency within the national basic research project ``Slovene scientific texts: resources and description'' (J6-7094, 2016--2019).

\bibliographystyle{splncs04}
\bibliography{xample}

\begin{thebibliography}{10}
\providecommand{\url}[1]{\texttt{#1}}
\providecommand{\urlprefix}{URL }
\providecommand{\doi}[1]{https://doi.org/#1}

\bibitem{bada2014concept}
Bada, M., Eckert, M., Evans, D., Garcia, K., Shipley, K., Sitnikov, D., Jr.,
  W.A.B., Cohen, K.B., Verspoor, K., Blake, J.A., Hunter, L.E.: Concept
  annotation in the {CRAFT} corpus. {BMC} Bioinformatics  \textbf{13}, ~161
  (2012)

\bibitem{Bernier-Colborne2014}
Bernier-Colborne, G., Drouin, P.: {Creating a test corpus for term extractors
  through term annotation}. Terminology  \textbf{20},  50--73 (2014)

\bibitem{conrado2013machine}
Conrado, M., Pardo, T., Rezende, S.: A machine learning approach to automatic
  term extraction using a rich feature set. In: Proceedings of the 2013 NAACL
  HLT Student Research Workshop. pp. 16--23 (2013)

\bibitem{kas-jtdh}
Erjavec, T., Fi\v{s}er, D., Ljube{\v s}i{\'c}, N., Logar, N., Ojster\v{s}ek,
  M.: Slovenska znanstvena besedila: prototipni korpus in načrt analiz
  (slovene scientific texts: Prototype corpus and research plan. In:
  Proceedings of the Conference on Language Technologies and Digital
  Humanities. Ljubljana University Press, Faculty of Arts (2016)

\bibitem{11356/1198}
Erjavec, T., Fi{\v s}er, D., Ljube{\v s}i{\'c}, N., Arhar~Holdt, {\\v S}.,
  Bren, U., Robnik~{\\v S}ikonja, M., Udovi{\v c}, B.: Terminology
  identification dataset {KAS}-term 1.0 (2018),
  \url{http://hdl.handle.net/11356/1198}, slovenian language resource
  repository {CLARIN}.{SI}

\bibitem{raslan-term}
Fi\v{s}er, D., Suchomel, V., Jakubi\v{c}ek, M.: Terminology extraction for
  academic slovene using sketch engine. In: RASLAN 2016 : Recent Advances in
  Slavonic Natural Language Processing. pp. 135--141 (2016)

\bibitem{Frantzi2000}
Frantzi, K., Ananiadou, S., Mima, H.: Automatic recognition of multi-word
  terms:. the c-value/nc-value method. International Journal on Digital
  Libraries  \textbf{3}(2),  115--130 (Aug 2000). \doi{10.1007/s007999900023},
  \url{https://doi.org/10.1007/s007999900023}

\bibitem{Nakagawa:Mori:03}
H., N., T., M.: Automatic term recognition based on statistics of compound
  nouns and their components. Terminology  \textbf{9}(2),  201--219 (2003)

\bibitem{handschuh2014acl}
Handschuh, S., QasemiZadeh, B.: The acl rd-tec: a dataset for benchmarking
  terminology extraction and classification in computational linguistics. In:
  COLING 2014: 4th International Workshop on Computational Terminology (2014)

\bibitem{kim2003genia}
Kim, J.D., Ohta, T., Tateisi, Y., ichi Tsujii, J.: {GENIA} corpus - a
  semantically annotated corpus for bio-textmining. In: ISMB (Supplement of
  Bioinformatics). pp. 180--182 (2003)

\bibitem{loukachevitch2012automatic}
Loukachevitch, N.V.: Automatic term recognition needs multiple evidence. In:
  LREC. pp. 2401--2407 (2012)

\bibitem{openscience-ojster}
Ojsteršek, M., Kotar, M., Ferme, M., Hrovat, G., Borovič, M., Bregant, A.,
  Bezget, J., Brezovnik, J.: Vzpostavitev repozitorijev slovenskih univerz in
  nacionalnega portala odprte znanosti (the set-up of the repository of slovene
  universities and the national portal of open science). Knjižnica
  \textbf{58}(3) (2014)

\bibitem{pazienza2005terminology}
Pazienza, M., Pennacchiotti, M., Zanzotto, F.: Terminology extraction: an
  analysis of linguistic and statistical approaches. Knowledge mining pp.
  255--279 (2005)

\bibitem{scikit-learn}
Pedregosa, F., Varoquaux, G., Gramfort, A., Michel, V., Thirion, B., Grisel,
  O., Blondel, M., Prettenhofer, P., Weiss, R., Dubourg, V., Vanderplas, J.,
  Passos, A., Cournapeau, D., Brucher, M., Perrot, M., Duchesnay, E.:
  Scikit-learn: Machine learning in {P}ython. Journal of Machine Learning
  Research  \textbf{12},  2825--2830 (2011)

\bibitem{pinnis12-term}
Pinnis, M., Ljube{\v s}i{\'c}, N., {\c S}tef{\u a}nescu, D., Skadi{\c n}a, I.,
  Tadi{\'c}, M., Gornostay, T.: Term extraction, tagging, and mapping tools for
  under-resourced languages. In: Proceedings of the Terminology and Knowledge
  Engineering (TKE2012) Conference (2012)

\bibitem{schafer2015evaluating}
Sch{\"a}fer, J., R{\"o}siger, I., Heid, U., Dorna, M.: Evaluating noise
  reduction strategies for terminology extraction. In: TIA. pp. 123--131 (2015)

\end{thebibliography}


\end{document}